\documentclass[11pt,a4paper]{article}
\usepackage{hyperref}
\usepackage[hyperref]{emnlp-ijcnlp-2019}
\usepackage{times}
\usepackage{latexsym}
\usepackage{booktabs} 
\usepackage{multirow}
\usepackage{url}
\usepackage{amsmath}
\usepackage{amsfonts}
\usepackage{graphicx}
\usepackage{bm}
\usepackage{color}
\usepackage{diagbox}
\usepackage{xspace}
\usepackage{courier}

\usepackage{bm}

\newcommand\OurModel{KGNN\xspace}

\usepackage{url}

\aclfinalcopy 

\title{Multi-Paragraph Reasoning with Knowledge-enhanced \\
Graph Neural Network}

\author{Deming Ye$^{1}$, Yankai Lin$^{2}$, Zhenghao Liu$^{1}$,  \textbf{Zhiyuan Liu$^{1}$, Maosong Sun$^{1}$}\\
$^{1}$Department of Computer Science and Technology, Tsinghua University, Beijing, China\\
Institute for Artificial Intelligence, Tsinghua University, Beijing, China\\
State Key Lab on Intelligent Technology and Systems, Tsinghua University, Beijing, China \\
$^{2}$Pattern Recognition Center, WeChat AI, Tencent Inc.\\
\texttt{ydm18@mails.tsinghua.edu.cn}
}

\date{}

\begin{document}
\maketitle
\begin{abstract}
Multi-paragraph reasoning is indispensable for open-domain question answering (OpenQA), which receives less attention in the current OpenQA systems. In this work, we propose a knowledge-enhanced graph neural network (KGNN), which performs reasoning over multiple paragraphs with entities. To explicitly capture the entities' relatedness, KGNN utilizes relational facts in knowledge graph to build the entity graph. The experimental results show that KGNN outperforms in both distractor and full wiki settings than baselines methods on HotpotQA dataset. And our further analysis illustrates KGNN is effective and robust with more retrieved paragraphs.

\end{abstract}

\section{Introduction}
\label{sec:introduction}

Open-domain question answering (OpenQA) aims to answer questions based on large-scale knowledge source, such as an unlabelled corpus. Recent years, OpenQA has aroused the interest of many researchers, with the availability of large-scale datasets such as Quasar~\citep{dhingra2017quasar}, SearchQA~\citep{dunn2017searchqa}, TriviaQA~\citep{joshi-EtAl:2017:Long}, etc. They proposed lots of OpenQA models~\citep{chen-EtAl:2017:Long4,clark2017simple,wang2018r3,wang2017evidence,choi2017coarse,lin2018denoising} which achieved promising results in various public benchmarks.

\begin{figure}[t]
    \centering
    \includegraphics[width=\columnwidth]{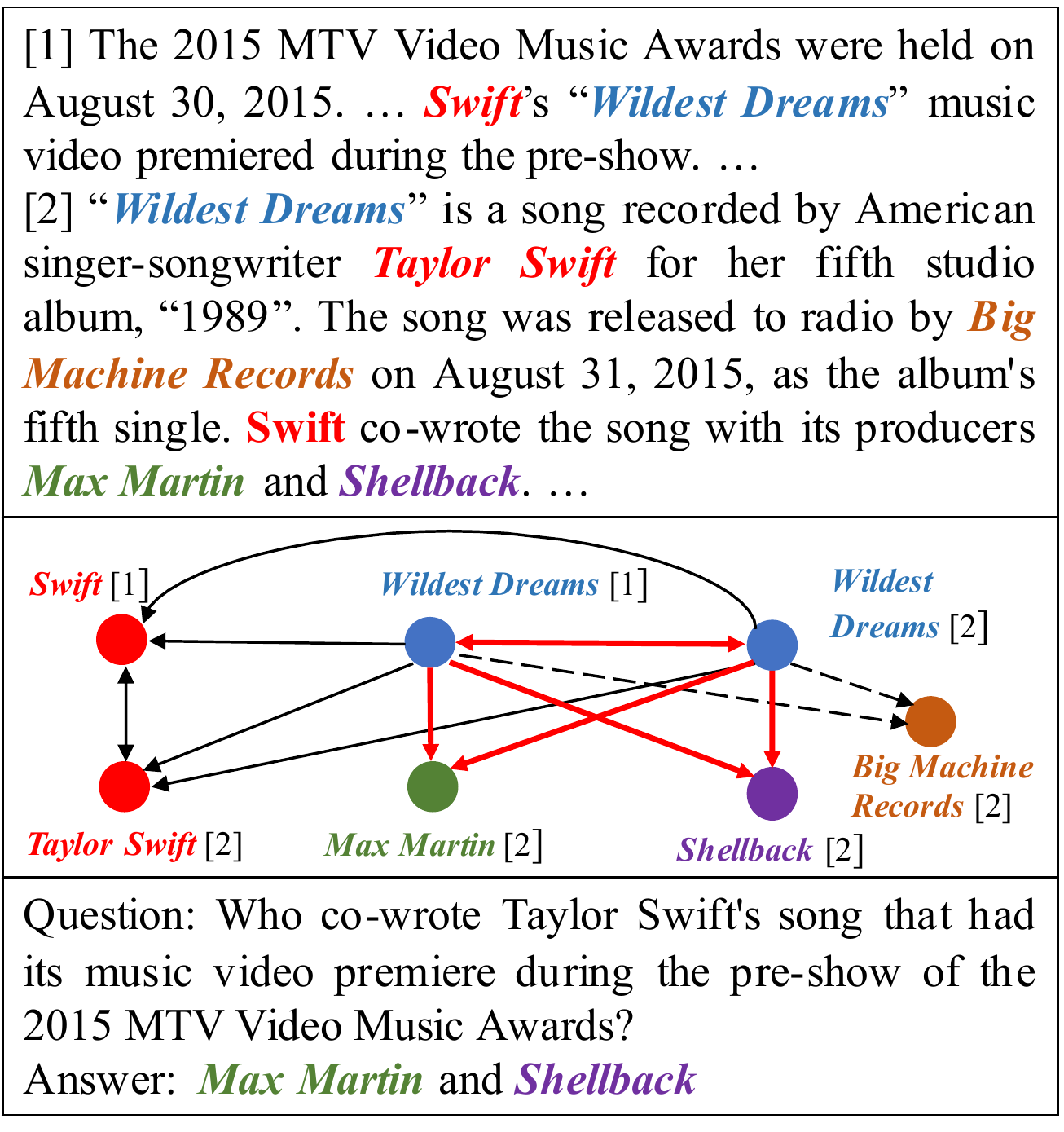}
    \caption{An example of multi-hop reasoning. \OurModel extracts entity from each paragraph, and propagates message through double-headed arrow \emph{equal\_to}, solid \emph{lyrics\_by}, and dotted \emph{record\_label} edges.  Edges helping us reasoning for the question are picked out in red. }
    \label{fig:example}
    \vspace{-1.5em}
\end{figure}
However, most questions in previous OpenQA datasets only require reasoning within a single paragraph or a single-hop over paragraphs. The HotpotQA dataset~\citep{yang2018hotpotqa} was constructed to facilitate the development of OpenQA system in handling multi-paragraph reasoning. Multi-paragraph reasoning is an important and practical problem towards a more intelligent OpenQA. Nevertheless, existing OpenQA systems have not paid enough attention to multi-paragraph reasoning. They generally fall into two categories when dealing with multiple paragraphs: (1) regarding each paragraph as an individual which cannot reason over paragraphs; (2) concatenating all paragraphs into a single long text which leads to time and memory consuming.

To achieve a multi-paragraph reasoning system, we propose a knowledge-enhanced graph neural network (KGNN). First, we build an entity graph by all named entities from paragraphs, and add co-reference edges to the graph if the entity appears in different paragraphs. After that, to explicitly capture the entities' relatedness, we further utilize the relational facts in knowledge graph (KG) to build the relational entity graph for reasoning, i.e, add a relation edge to the graph if two entities have a relation in KG. We believe that the reasoning information can be captured through propagation over a relational entity graph.
As the example in Figure \ref{fig:example}, for the given entity \emph{Wildest Dreams}, we require two kinds of the one-hop reasoning to obtain the answer: One is \emph{Wildest Dreams} appears in multi-paragraph and the other is reasoning based on the relational fact (\emph{Wildest Dreams},\emph{lyrics by}, \emph{Max martin} and \emph{Shellback}).

Our main contribution is that we propose a novel reasoning module combined with knowledge. The experiments show that reasoning over entities can help our model surpass all baseline models significantly on HotpotQA dataset. Our analysis demonstrates that \OurModel is robust and has a strong ability to handle massive texts.






\section{Model architecture}
\label{sec:method}

In this section, we introduce the framework of knowledge-enhanced graph neural network (KGNN) for multi-paragraph reasoning.

\begin{figure}[t]
    \centering
    \includegraphics[width=\columnwidth]{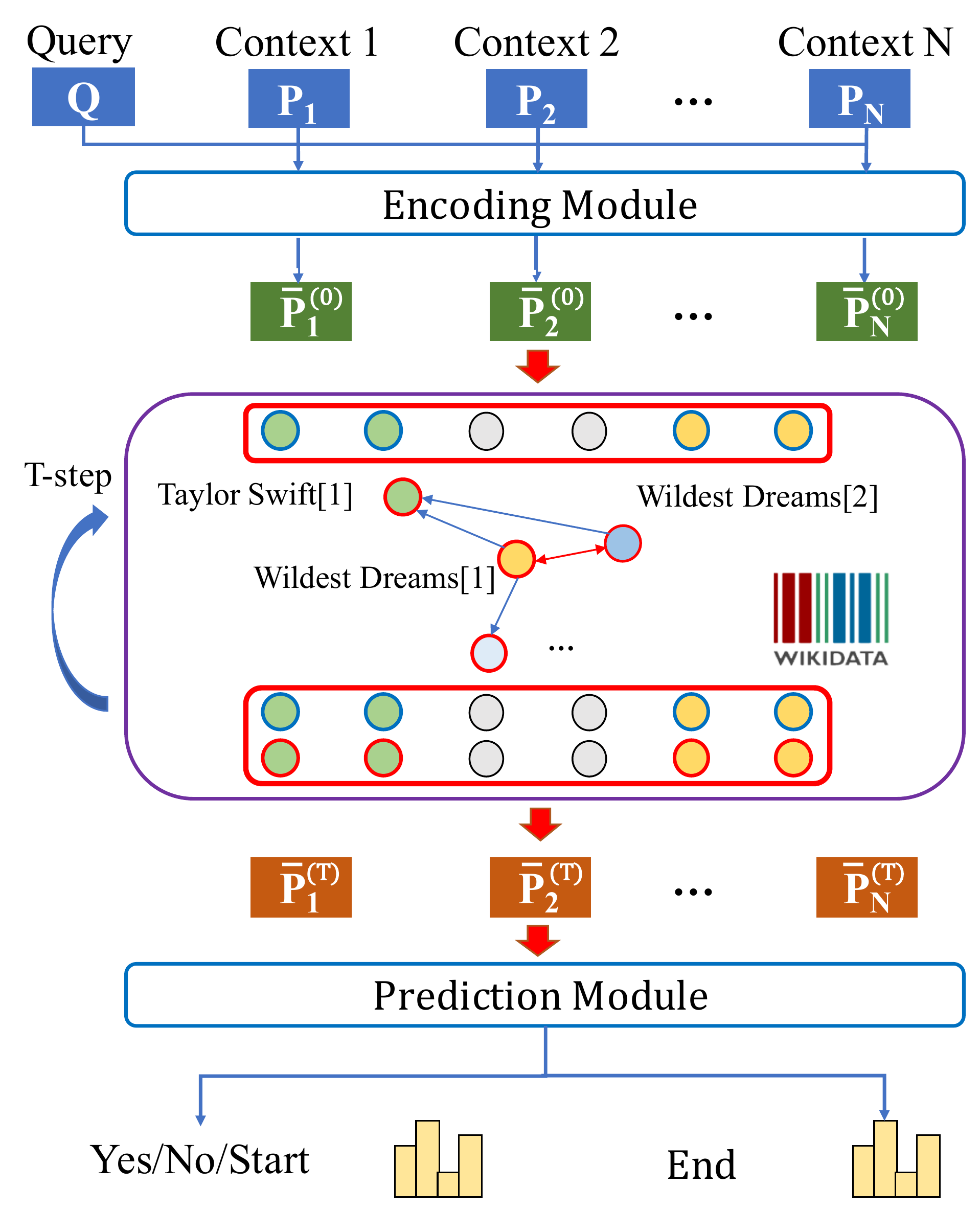}
    \caption{Overview of \OurModel.}
    \label{fig:main}
    \vspace{-1.5em}
\end{figure}

As shown in Figure~\ref{fig:main}, \OurModel consists of three parts including encoding module, reasoning module, and prediction module. An additional component is used for supporting fact prediction.

\subsection{Encoder module}


Without loss of generality, we use the encoding components described in~\citet{clark2017simple}, which include a character-level encoder, self-attention layer, and bi-attention layer to embed the question and paragraphs into their low-dimensional representations. The question $Q$ and paragraphs $\{P_1, P_2, \cdots, P_N\}$ are first encoded as:
\begin{eqnarray}
    \bm{Q} &=& \texttt{Self-Att}(\texttt{Char-Enc}(Q)),\\
    \bm{P}_i &=& \texttt{Self-Att}(\texttt{Char-Enc}(P_i)),
\end{eqnarray}
then we compute the question-related paragraph representations through a bi-attention operation:
\begin{equation}
    \bar{\bm{P}_i}^{(0)} = \texttt{Bi-Att}(\bm{Q},\bm{P}_i),
\end{equation}
where $\texttt{Char-Enc}(\cdot)$ denotes the ``character-level encoder'', $\texttt{Self-Att}(\cdot)$ denotes the ``self-attention layer'' and $\texttt{Bi-Att}(\cdot)$ denotes the ``bi-attention layer''. 

\subsection{Reasoning module}

Most existing OpenQA models~\citep{chen-EtAl:2017:Long4, clark2017simple}  directly regards all paragraphs as individuals, which may have obstacles in multi-paragraph reasoning for answering questions. Different from existing OpenQA models, \OurModel propagates information among paragraphs with entities through a knowledge-enhanced graph neural network.

\subsubsection{Build entity graph}

We regard all entities\footnote{We link the named entities to Wikidata with spaCy (\url{https://spacy.io/})} from paragraphs as nodes in the graph for multi-paragraph reasoning, which is denoted by $\bm{V}$. And we build the entity graph according to two types of edges among paragraphs.


Co-reference of entity is the most common relation across paragraphs. For two nodes $v_i, v_j\in \bm{V}$, an edge ${\hat{e}}_{ij}=(v_i, v_j)$ will be added to the graph if two nodes indicate the same entity; Furthermore, we adopt relational facts from knowledge graph to enhance our model's reasoning ability. For two nodes $v_i, v_j\in \bm{V}$, a relational edge ${{\bar{e}}_{ij}}^{\,\text{r}}=(v_i, v_j)$ will be added to the graph if two entities have a relation $r$, which could help the relational reasoning.

\subsubsection{Relational Reasoning} 
As we have built the graph $\mathcal{G}=(\bm{V},\bm{E})$, we leverage \OurModel to perform reasoning. The reasoning process is as follows:


\textbf{Graph Representation.} For each node $v_i\in \bm{V}$, we obtain its initial representation from contextual word representations. Assuming the corresponding entity has $k$ mentions in the paragraph, the initial node representation $\mathbf{v}_i$ would be defined as:
\begin{equation}
\label{eq:reason_start}
\mathbf{v}_i = \texttt{Max-Pool}(\mathbf{m}_1,\mathbf{m}_2,..\mathbf{m}_k),
\end{equation}
where $\texttt{Max-Pool}(\cdot)$ denotes a max-pooling layer and $\mathbf{m}_i$ denotes the representation of the mention $m_i$. Here, if a mention $m_i$ ranges from $s$-th to $t$-th word in the paragraph $P_j$, its representation is defined as the mean of word representations as $\mathbf{m}_i = \frac{1}{t-s+1}{\sum_{l=s}^t \texttt{FFN}(\mathbf{\bar{P}}_{jl})}$, where $\texttt{FFN}(\cdot)$ indicates a fully-connect feed-forward layer. 






\textbf{Message Propagation.}
As we want to reason over paragraphs with entities, we propagate messages from each node to its neighbors to help perform reasoning. Since different kinds of edges play different roles in reasoning, we use the relation-specific network in the message propagation.
Formally, we define the following propagation function for calculating the update of a node:
\begin{equation}
\mathbf{v}_i^{u} = \sum_r \frac{\alpha_r}{|N_r(v_i)|} \sum_{v_j \in N_r(v_i)}  \phi_r(\mathbf{v}_j),
\end{equation}
where $N_r(v_i)$ is the neighbor set for $v_i$ with relation $r$. $\alpha_r$ and $\phi_r(\cdot)$ are relation-specific attention weight and feed-forward network respectively, which is defined as follows.

To measure the relevance between question and relation, we utilize a entire question representation $\bar{\bm{Q}}$ to compute relation-specific attention weight for relation $r$ as $\mathbf{\alpha}_r = \texttt{softmax}(\texttt{FFN}(\mathbf{\bar{\bm{Q}}})\big)$. And with a translating relation embedding $\mathbf{E}_r$, we design our relation-specific network as $\phi_r(\mathbf{v}_j) = \texttt{FFN}(\mathbf{v}_j + \mathbf{E}_r )$.



\textbf{Paragraph Update.} After message propagation over entities among paragraphs, we update the question-aware paragraph representations. We first define question-aware paragraph representations $\mathbf{U}_{ij}$ is defined as:
\begin{equation}
\mathbf{U}_{ij} = 
    \begin{cases}
    \bm{v}_{Idx(ij)}^u &  \text{an entity appears at $P_{ij}$} , \\
    \bm{0}  & \text{otherwise},
    \end{cases}
\end{equation}
where $Idx(\cdot)$ indicates the node index of $P_{ij}$ in the constructed entity graph. 

Further, we utilize a reset gate to decide how much information to keep from the graph-aware paragraph representation:
\begin{align}
\mathbf{r}_{ij} &= \sigma ( \mathbf{W}^p\mathbf{\bar{P}}_{ij} + \mathbf{W}^u \mathbf{U}_{ij}), \\
\mathbf{\hat{U}}_{ij} &= \mathbf{r}_{ij} * \mathbf{U}_{ij}+ (1-\mathbf{r}_{ij})*\mathbf{\bar{P}}_{ij},
\end{align}
where $\mathbf{W}^p$ and $\mathbf{W}^u$ are trainable matrices and $\sigma$ denotes the sigmoid function.

Finally, we apply self-attention mechanism to share global information from entities to the whole paragraph, and the output of reasoning part will be added to the input as a residual:
\begin{equation}
\label{eq:reason_end}
\mathbf{\bar{P}}_{ij}' = \mathbf{\bar{P}}_{ij} + \texttt{Self-Att}(\hat{\mathbf{U}}_{ij}).
\end{equation}

We denote the initial paragraph representations as $\mathbf{\bar{P}}^{(0)}$, and denote the entire one-step reasoning process, i.e., Eq. (\ref{eq:reason_start}-\ref{eq:reason_end}) as a single function:
\begin{equation}
    \mathbf{\bar{P}}^{(t)} = \texttt{Reason}(\mathbf{\bar{P}})^{(t-1)},
\end{equation}
where $t\ge 1$. Hence, a $T$-step reasoning can be divided into  $T$ times one-step reasoning.

\subsection{Prediction module}

After $T$-step reasoning on the relational entity graph, we predict the answer according to the final question-aware paragraph representations $\mathbf{\bar{P}}^{(T)}$. Furthermore, for the sake of the answer probability comparison among multiple paragraphs, we utilize shared-normalization \citep{clark2017simple} in the answer prediction. 




\section{Experiment}
\subsection{Dataset}
We use HotpotQA to conduct our experiments. HotpotQA is an OpenQA dataset with complex reasoning, which contains more than $11k$ question-answer pairs and all questions in development and testing set require complex multi-hop reasoning over paragraphs.
\begin{table*}
\centering
\small

\begin{tabular}{l |l|c c |c c| c c}
\toprule

\multirow{2}{*}{Model}  & \multirow{2}{*}{Setting}  &\multicolumn{2}{c|}{Ans}   &  \multicolumn{2}{c|}{Sup Fact}    & \multicolumn{2}{c}{Joint} \\
                        &                           & EM        & F1            & EM        & F1                    & EM        & F1\\
\midrule
\citet{yang2018hotpotqa} & distractor                & 45.60     & 59.02         & 20.32     & 64.49                 & 10.83     &40.16 \\
KGNN           & distractor                &\bf{50.81} &\bf{65.75}     & \bf{38.74}& \bf{76.79}            & \bf{22.40}& \bf{52.82}\\
\midrule
\citet{yang2018hotpotqa} & full wiki                  & 23.95     & 32.89         & 3.86      &37.71                  & 1.85      & 16.15\\
KGNN           & full wiki                  &\bf{27.65} &\bf{37.19}     & \bf{12.65}& \bf{47.19}            & \bf{7.03} &\bf{24.66}\\

\bottomrule
\end{tabular}
\caption{Results on HotpotQA test set for \emph{distractor} and \emph{full wiki} settings.}
\label{tab:overall_result}
\end{table*}

\begin{table}
\centering
\small

\begin{tabular}{l|c c }
\toprule
Model             & EM        & F1\\
\midrule
\citet{yang2018hotpotqa}-split      & 11.87     & 41.87\\
\citet{yang2018hotpotqa}      & 18.14     & 50.72\\
KGNN (\#Layer=1)                    & 22.26     & 53.50\\
KGNN (\#Layer=2)                    &\textbf{22.41}& \textbf{54.05}\\
KGNN (\#Layer=3)                    & 22.24     & 53.49\\
\bottomrule
\end{tabular}
\caption{Effect of Layer Number on joint metrics.}
\label{tab:layer_number}
\end{table}

We evaluate \OurModel on \emph{distractor} and \emph{full wiki} settings of HotpotQA. In \emph{distractor} setting, $2$ golden paragraphs are given as inputs with $8$ disturbance terms. In \emph{full wiki} setting, only questions are offered and we need to extract the answer from the whole corpus. So we employ an information retrieval system to retrieve $30$ paragraphs from the entire Wikipedia for the following experiment.

For the relational entity graph construction, we align entities to Wikidata items and we add $5$ common relations in question answering to propagate message. The relations include \{\emph{director}, \emph{position\_held}, \emph{record\_label}, \emph{lyrics\_by}, \emph{adapted\_from} \}.

\subsection{Baseline Methods}
To verify the effectiveness of our reasoning module, we compare \OurModel with \citet{yang2018hotpotqa}\footnote{We use Adam optimizer to train the official code and gain 10\% joint F1 promotion} and a modified version, named \citet{yang2018hotpotqa}-split, which regards all paragraphs individually and applies a shared-normalization function over paragraphs to obtain the answer.

\subsection{Overall Result}

Table \ref{tab:overall_result} shows experimental results on HotpotQA dataset. The table illustrates that our model outperforms \citet{yang2018hotpotqa} in both \emph{distractor} and \emph{full wiki} settings. It indicates the effectiveness of KGNN model in multi-paragraph reasoning with external knowledge.

\subsection{Effect of Layer Number}

In this part, we additionally perform experiments to understand the effect of the layer number on \OurModel. From Table \ref{tab:layer_number}, we could observe that $2$-layer \OurModel model achieves the best performance. From carefully analyzing the data, we find that most questions in HophotQA dataset only require $2$-hop reasoning for obtaining the answer. And 2-layer KGNN model has the ability to capture enough reasoning flows in this scenario.


\subsection{Effect of Paragraph Number}

In real world (i.e., \emph{full wiki} setting), an OpenQA system usually covers enough paragraphs to provide useful information for answering questions. Nevertheless, the noise of paragraphs misleads models and remains a challenge for OpenQA.


For each question, we use $30$ retrieved paragraphs to investigate the denoising ability of the system. As shown in Table \ref{tab:paragraph_num}, \OurModel model handles and acquires more information from extra paragraphs. \OurModel's performance increases more significantly than \citet{yang2018hotpotqa}. The results demonstrate that reasoning across paragraphs through entities and their relations is a robust and flexible way of encoding multiple paragraphs.

\begin{table}[t]
\centering
\small
\begin{tabular}{l |c| c | c}
\toprule
Model & 10  & 20& 30\\
\midrule
\citet{yang2018hotpotqa}  & 20.77& 20.88 &21.52\\
KGNN & \textbf{22.20} &  \textbf{25.02} & \textbf{25.12}\\
\bottomrule
\end{tabular}
\caption{Effect of paragraph number on joint F1.}
\label{tab:paragraph_num}
\end{table}

\section{Related Work}
\label{sec:related_work}
Open-domain question answering (OpenQA) aims to answer the given question by leveraging knowledge sources, which is first proposed by \citet{green1961baseball}. Recently, \citet{chen-EtAl:2017:Long4} introduces machine reading comprehension technique to answer open-domain questions by reading multiple retrieved texts. \citet{wang2018r3} selects a most informative paragraph to answer the question. Moreover, \citet{wang2017evidence}, \citet{lin2018denoising} and \citet{clark2017simple} focus on how to aggregate evidence in multiple paragraphs. However, all existing OpenQA models simply regard each paragraph as an individual or concatenate all paragraphs into a single long text.

\citet{Haitian2019open}, \citet{de2018question} and \citet{cao2019bag} introduce entity graph neural network to reason over entities for Knowledge Base QA (KBQA). Different from KBQA, our work regards entities as bridges to reason over paragraphs and can give a span-style answer instead of supplying an entity-style answer.

\section{Conclusion}

Multi-paragraph reasoning is crucial for answering open-domain questions in practice, while it is still not considered in most existing OpenQA systems. In this work, we propose a novel OpenQA model \OurModel, which performs reasoning over paragraphs via a knowledge enhanced graph neural network. Experimental results show that \OurModel  outperforms strong baselines with a large margin on the HotpotQA dataset, and also has the ability to tackle more informative texts. We hope our work can shed some lights to the combination of knowledge graph and text for OpenQA.


\bibliography{emnlp-ijcnlp-2019}
\bibliographystyle{acl_natbib}

\appendix

\label{sec:supplemental}

\end{document}